\documentclass[lettersize,journal]{IEEEtran}
\usepackage{amsmath,amsfonts}
\usepackage{algorithmic}
\usepackage{algorithm}
\usepackage{array}
\usepackage{booktabs}
\usepackage[caption=false,font=normalsize,labelfont=sf,textfont=sf]{subfig}
\usepackage{textcomp}
\usepackage{stfloats}
\usepackage{multirow}
\usepackage{url}
\usepackage{verbatim}
\usepackage{graphicx}
\usepackage{amssymb}
\usepackage{booktabs}
\usepackage{amssymb}
\usepackage{bm}
\usepackage{makecell}
\usepackage[table]{xcolor}
\usepackage[colorlinks,linkcolor=blue]{hyperref}
\usepackage{rotating}
\usepackage{ulem}

\begin{document}
\title{SWA-PF: Semantic-Weighted Adaptive Particle Filter for Memory-Efficient 4-DoF UAV Localization in GNSS-Denied Environments}

\author{
Jiayu Yuan,
Ming Dai,
Enhui Zheng,
Chao Su,
Nanxing Chen,
Qiming Hu,
Shibo Zhu and 
Yibin Cao
\IEEEcompsocitemizethanks{
J. Yuan, E. Zheng, C. Su, N. Chen, Q. Hu, S. Zhu and Y. Cao are with the School of China Jiliang University, Hangzhou, 310018, China. Email: (p23010854149, ehzheng, p24010854093, p23010854013, p23010854043, p24010854166, 2200602217)@cjlu.edu.cn. 

\IEEEcompsocthanksitem M. Dai is with the School of Automation, Southeast University, Nanjing 210096, China. E-mails: mingdai@seu.edu.cn. 
}
}

\maketitle

\begin{abstract}
Vision-based Unmanned Aerial Vehicle (UAV) localization systems have been extensively investigated for Global Navigation Satellite System (GNSS)-denied environments.
However, existing retrieval-based approaches face limitations in dataset availability and persistent challenges including suboptimal real-time performance, environmental sensitivity, and limited generalization capability, particularly in dynamic or temporally varying environments.
To overcome these limitations, we present a large-scale Multi-Altitude Flight Segments dataset (MAFS) for variable altitude scenarios and propose a novel Semantic-Weighted Adaptive Particle Filter (SWA-PF) method.
This approach integrates robust semantic features from both UAV-captured images and satellite imagery through two key innovations:
a semantic weighting mechanism and an optimized particle filtering architecture.
Evaluated using our dataset, the proposed method achieves 10× computational efficiency gain over feature extraction methods, maintains global positioning errors below 10 meters, and enables rapid 4 degree of freedom (4-DoF) pose estimation within seconds using accessible low-resolution satellite maps.
Code and dataset will be available at https://github.com/YuanJiayuuu/SWA-PF.
\end{abstract}

\begin{IEEEkeywords}
Localization, UAV, Semantic segmentation, Scene understanding.
\end{IEEEkeywords}

\section{Introduction}\label{sec1}

The rapid advancement of unmanned aerial technology has propelled Unmanned Aerial Vehicles (UAVs) to prominence across various domains, including environmental exploration \cite{xu2023vision}, precision agriculture \cite{bib1}, terrain surveillance \cite{bib2}, and search-and-rescue operations \cite{rescue_tasks2}. At present, the most widely used localization method for UAVs is the global navigation satellite system (GNSS) \cite{GNSS}. However, GNSS signals are highly susceptible to interference from obstacles and complex electromagnetic environments \cite{bib222}. Consequently, developing real-time autonomous navigation and localization technologies for UAVs is of significant practical importance \cite{review_new}.

Vision-based localization, as an alternative autonomous method, has gained increasing attention in recent years \cite{review}. This technique computes spatial coordinates by analyzing UAV-captured top-view images and aligning them with corresponding satellite images. Its immunity to electromagnetic interference makes it particularly effective in extreme environments where traditional positioning systems fail, offering a robust supplementary solution to conventional geolocation methods \cite{DenseUAV}.

\begin{figure}[!t]
    \centering
    \includegraphics[width=0.95\linewidth]{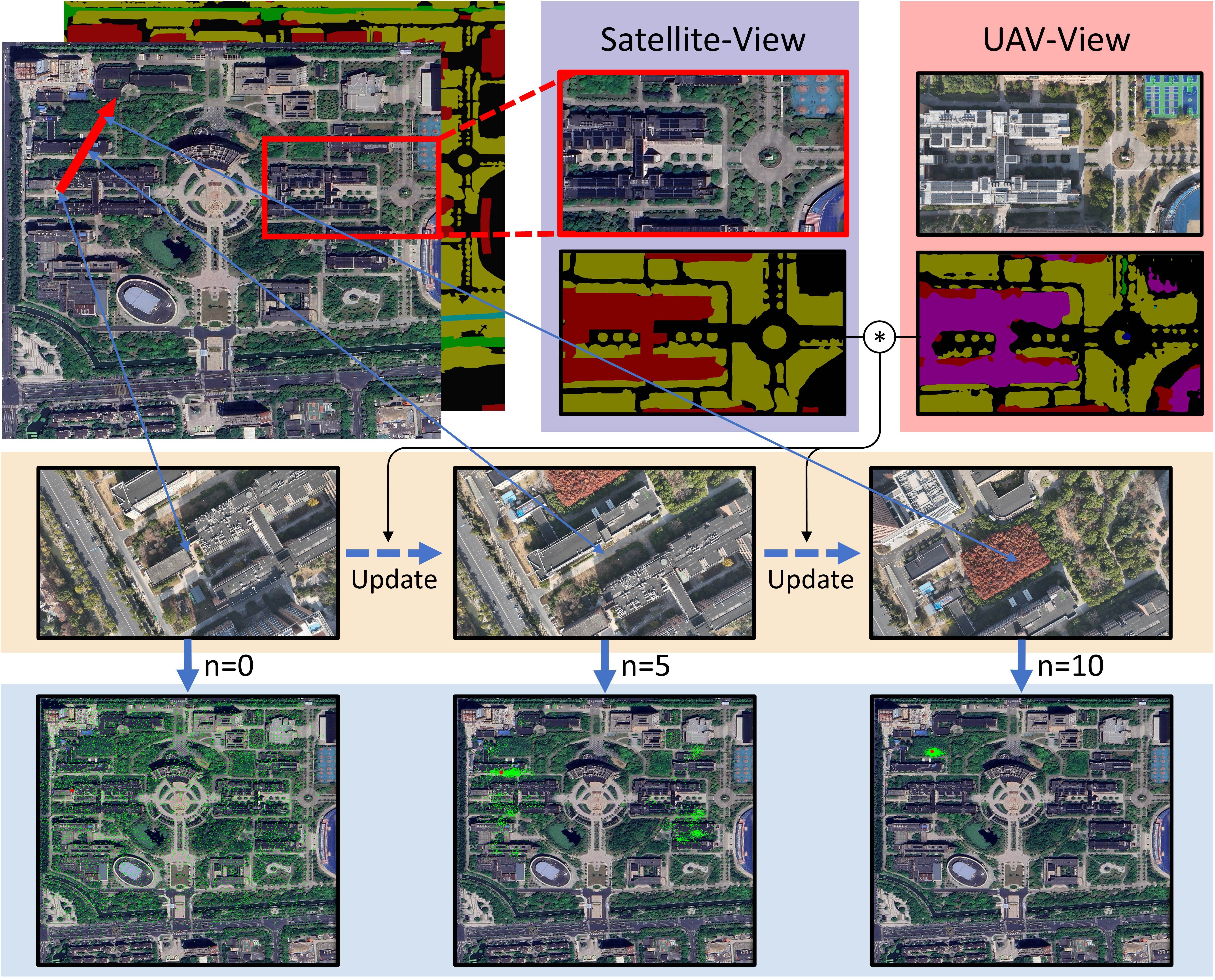}
    \caption{The simplified pipeline of the proposed visual positioning method. The upper part shows the UAV-view, the satellite-view, and their corresponding semantic images. While cross-view images exhibit substantial visual differences, they retain comparable semantic features. The lower part illustrates the iterative process of UAV visual positioning using semantic similarity within a particle filter framework.}
    \label{fig1}
\end{figure}

Existing UAV localization datasets \cite{UL14, CVUSA, SUES, 1652} commonly adopt a one-to-one matching strategy, where UAV top-view images are paired with corresponding satellite image patches for localization. However, this approach faces three main limitations: (1) discontinuous sampling patterns, as seen in \cite{1652}; (2) fixed-altitude data collection, such as in \cite{DenseUAV}, which limits its application in variable-altitude scenarios; and (3) a lack of diversity in samples due to the reliance on paired data structures.

To address these issues, we introduce the Multi-Altitude Flight Segments (MAFS) dataset, a novel resource designed to provide high-quality, application-driven data for UAV localization research. This dataset includes authentic aerial footage captured across 14 university campuses in Hangzhou, China. The flight trajectories encompass both fixed- and variable-altitude configurations, covering a wide range of urban environments such as roads, buildings, and parks. Each video frame is accompanied by synchronized IMU sensor data, offering precise geospatial coordinates, altitude, yaw angle, and velocity. Additionally, the positioning target of each segment is matched with a wide-area satellite image, covering a search range of up to 2 km × 2 km. In response to the growing trend of using semantic scene understanding for localization, we also introduce the SemanticMAFS dataset, a semantic segmentation dataset derived from the MAFS dataset for UAV top-view images.

Current absolute visual localization methods for UAVs can be categorized into four main paradigms: 
(1) Template matching \cite{bib17, UAVD4L}, which determines position by pixel-level comparison between UAV images and reference satellite maps, though it requires strict visual similarity between the two. 
(2) Feature point matching \cite{bib20,abb} identifies positional relationships through key visual elements, but its reliance on extensive feature extraction and matching makes it computationally intensive, limiting its real-time applicability. 
(3) Visual odometry \cite{bib24,SLAM1} reconstructs relative trajectories by estimating motion between consecutive frames, but the accumulated drift over time degrades accuracy during extended operations.
(4) Deep learning methods \cite{bib25,OSFPI,bib27,UL14} leverage neural networks for end-to-end localization via image feature extraction, but their performance degrades in unseen environments due to limited generalization.

Feature-based visual localization methods, when directly applied to satellite imagery, prove inefficient. Satellite images are often captured at varying temporal intervals, necessitating UAV-based visual localization systems that can account for seasonal variations \cite{season1, season2}. Moreover, these systems must demonstrate robust applicability across diverse geographical regions, as the heterogeneity of multisource images \cite{bib301} poses significant challenges for image matching and localization.

To overcome these challenges, we propose the Semantic-Weighted Adaptive Particle Filter (SWA-PF), a novel retrieval framework that integrates semantic segmentation to extract critical features from both UAV-captured and satellite images \cite{sec_review}. Recent advancements in semantic-aware approaches have shown promising capabilities in addressing variations in perspective and differences across sensing modalities \cite{bib12}. As depicted in Fig. \hyperref[fig1]{1}, inspired by the success of Monte Carlo methods in robot localization \cite{Mc1}, we combine semantic segmentation techniques from computer vision \cite{sec_review, VDD, UAVid} with the Monte Carlo localization framework to handle environmental changes and enhance the generality of localization methods. We optimized the initialization, update, and clustering procedures of the particle filter, achieving real-time computational efficiency (within seconds).

To validate the effectiveness of SWA-PF, we conducted extensive experiments on the MAFS dataset. Experimental results demonstrate that SWA-PF achieves positioning accuracy within 10 meters while offering superior computational efficiency and robustness compared to existing approaches. Our analysis confirms that the semantic-weighted mechanism enhances localization accuracy while optimizing operational efficiency. These findings establish SWA-PF as a viable solution for robust UAV navigation in dynamic, real-world environments.

The contributions of this work are as follows:
\begin{itemize}
    \item{We introduce a real-time cross-view UAV localization framework, Semantic-Weighted Adaptive Particle Filter (SWA-PF), which uses UAV images to localize a low-precision map. Our method not only determines latitude and longitude but also estimates the UAV's altitude and heading, achieving 4-DoF localization.}
    \item{We develop a new benchmark dataset, MAFS, based on the SWA-PF framework.}
    \item{We further annotate the MAFS dataset to create the SemanticMAFS dataset, which facilitates future UAV top-view image semantic segmentation applications.}
    \item{We release both our dataset and localization code, enabling the broader research community to extend and build upon our work.}
\end{itemize}

\section{Related Works}
\label{section II}
This section first reviews the geographical cross-view localization datasets in Section \hyperref[sectionIIA]{IIA}, followed by a discussion of existing absolute UAV visual localization methods in Section \hyperref[sectionIIB]{IIB}. Finally, Section \hyperref[sectionIIC]{IIC} introduces the important role of semantic understanding in robot system positioning.

\subsection{Comparison of Cross-View Localization Datasets}
\label{sectionIIA}
UAV localization datasets vary in terms of technical focus and application. The CVUSA dataset \cite{CVUSA} pairs panoramic ground-level images with corresponding satellite imagery from diverse regions across the United States, designed for one-to-one ground-to-satellite matching. It offers extensive geographic coverage and varied training samples for cross-view correlation learning. The VIGOR dataset \cite{Vigor} expands this concept by incorporating street-view images aligned with satellite tiles from a broader geographic area. Its innovative one-to-many matching framework improves model robustness, particularly in complex urban environments. The University-1652 dataset \cite{1652} introduces tri-view matching, combining ground, UAV, and satellite perspectives, with over 500,000 images of 1,652 campus buildings. This approach enables novel UAV navigation tasks while extending cross-view analysis to aerial platforms. SUE-200 \cite{SUES} specializes in altitude-variant analysis, with UAV imagery captured at four different flight levels paired with corresponding satellite data. This multi-altitude sampling enhances model adaptability across varying UAV operational heights. Finally, UL14 \cite{UL14} extends DenseUAV \cite{DenseUAV} by incorporating multi-temporal satellite imagery from 14 universities over two years. This dataset models UAV-satellite matching as a one-to-many problem, using dense spatial sampling and temporal variations to support GPS-denied self-localization under dynamic environmental conditions.

\subsection{Absolute Visual Localization Algorithm for UAVs}
\label{sectionIIB}

\subsubsection{Template Matching Method}
Template matching involves comparing live UAV images with reference satellite maps or georeferenced images through pixel-by-pixel analysis to calculate similarity measurements and identify spatial correspondences for precise localization. Dalen et al. \cite{bib17} proposed an absolute localization method for UAVs using image alignment and particle filtering, matching UAV images with a global reference map for accurate localization. Patel \cite{bib18} developed a visual localization framework for UAVs in outdoor GPS-denied environments, leveraging a particle filter and normalized information distance for position estimation. Wu et al. \cite{UAVD4L} introduced a UAV dataset and a two-stage localization pipeline, improving localization in GPS-denied environments with urban and rural scene coverage, accurate ground truth poses, and sensor data integration.

\subsubsection{Feature Point Matching Method}
Feature point matching identifies key visual elements in images, extracts their descriptors, and aligns features across multiple images to determine positional relationships. Several studies have advanced UAV position estimation using this approach. Saranya et al. \cite{bib20} proposed a method for UAV position estimation using RANSAC \cite{RANSAC} and SURF \cite{SURF} feature points, matching UAV images with a global reference map. Shan and Charan \cite{bib22} introduced a method for Google map-referenced UAV navigation, employing simultaneous feature detection and description. Mantelli et al. \cite{bib51} proposed a novel measurement model based on abBRIEF for global localization of UAVs over satellite images, utilizing a particle filter for accurate position estimation. In a follow-up study, Mantelli et al. \cite{abb} further developed this model, continuing their focus on global localization using abBRIEF.

\subsubsection{Visual Odometry Method}
Visual odometry estimates the motion between consecutive image frames to reconstruct a relative trajectory, integrating this with georeferenced data for absolute positioning. Greeff et al. \cite{bib24} explored a visual teach-and-repeat system that enables UAVs to autonomously return to their home position during GPS failure, utilizing visual information for route following and localization.

\subsubsection{Deep Learning Method}
Dai et al. \cite{bib25} introduced deep learning transformer architectures for feature extraction in image processing. They also proposed a multi-sampling strategy to address mismatches in satellite image sources and quantities. The following year, Dai et al. \cite{UL14} developed FPI, a cross-view geolocation method that diverges from traditional image retrieval approaches. This method uses a dual-branch network with non-shared weights to extract features from UAV and satellite images, generating a heatmap to predict the UAV location based on the highest response value. Recently, Chen et al. \cite{OSFPI} introduced OS-FPI, a coarse-to-fine single-stream network that significantly improves localization accuracy through offset prediction while minimizing network parameters. Wang et al. \cite{bib27} proposed Dynamic Weighted Decorrelation Regularization (DWDR) and a symmetrical sampling strategy to reduce feature embedding redundancy and enhance performance in cross-view geolocation tasks.

\subsection{Semantic-based Scene Localization Method}
\label{sectionIIC}
Recent studies have focused on semantic representations and matching for cross-view and cross-modal place recognition. Approaches such as semantic graph matching \cite{bib28}, semantic histograms \cite{bib29}, bag of words \cite{bib30}, and maximum clique \cite{bib31} have been explored. Additionally, methods leveraging geometric information of semantic objects, including shape, density, and contour, for position recognition have been proposed \cite{bib101}. Unlike these similarity-based methods, which compare semantics between two frames of images or point clouds, our approach focuses on the matching problem between small and extensive features. Hong et al. \cite{bib35} introduced a vision-based navigation approach using semantically segmented aerial images modeled as Gaussian mixture models, enabling efficient position estimation in GNSS-denied environments.

\section{Dataset}
\label{sectionIII}
This section introduces the UAV localization dataset MAFS and its corresponding semantic dataset, SemanticMAFS. We first highlight their distinctive characteristics compared with existing datasets, followed by a description of the sampling methodology and dataset structure.

\subsection{UAV Localization Dataset: MAFS}

In UAV localization research, the quality and representativeness of datasets are critical for reliable algorithm evaluation. We initially considered the UL14 dataset \cite{UL14} for testing. However, it exhibits several limitations, including limited altitude variation, relatively low image resolution, discontinuous flight trajectories, and constrained image dimensions. These drawbacks hinder fine-grained feature extraction and restrict accurate performance assessment. Although other datasets provide extensive coverage, most rely on fragmented sampling strategies (see Table \hyperref[tab1]{I}).  

\begin{table*}[!h]
    \caption{A summary of existing geo-localization datasets} 
    \label{tab1}
    \resizebox{\textwidth}{!}{
    \begin{tabular}{lllllll}
    \hline
    \textbf{Dataset} & \textbf{Altitude (m)} & \textbf{Platform} & \textbf{Target} & \textbf{Sampling} & \textbf{Source} & \textbf{Imgs./Platform}\\
    \hline
    UL14 \cite{UL14}            & 90--120   & Drone, Satellite         & UAV      & Discrete & Real Scenes & 3 + 6\\
    CVUSA \cite{CVUSA}          & -         & Drone, Ground, Satellite & User     & Discrete & Google Maps & 1 + 1\\
    SUES-200 \cite{SUES}        & 150--300  & Drone, Satellite         & Diverse  & Discrete & Google Maps & 50 + 1\\
    University-1652 \cite{1652} & -         & Drone, Ground, Satellite & Building & Discrete & Google Maps & 54 + 16.64 + 1\\
    \hline
    MAFS (Ours)                 & 100--500  & Drone, Satellite         & UAV      & Dense    & Real Scenes & \textgreater100 + 1\\
    \hline
    \end{tabular}}
\end{table*}

To overcome these limitations, we introduce the Multi-Altitude Flight Segments (MAFS) dataset, a large-scale UAV localization dataset tailored to real-world operational scenarios. MAFS offers high-resolution, dense, and multi-altitude data, thereby providing a more robust foundation for advancing aerial positioning research.

\begin{figure*}[!h]
    \centering
    \includegraphics[width=0.95\textwidth]{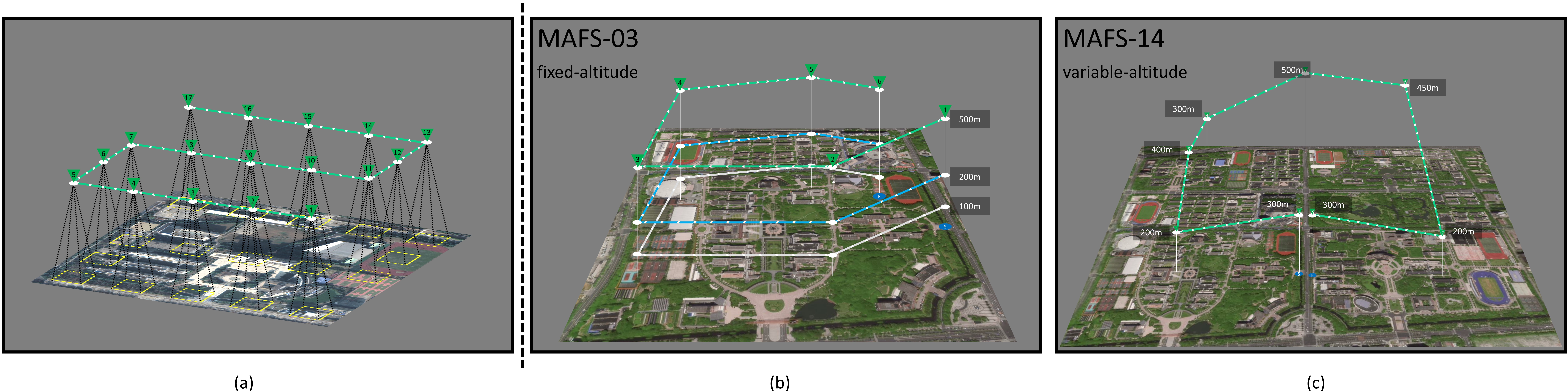}
    \caption{(a) Conventional retrieval-based datasets. The proposed MAFS dataset incorporates two flight strategies: (b) fixed-altitude configurations (100m, 200m, 300m, 400m, 500m) and (c) variable-altitude configurations (150--500m).}
    \label{fig2}
\end{figure*}

\begin{figure*}[!b]
    \centering
    \includegraphics[width=0.9\textwidth]{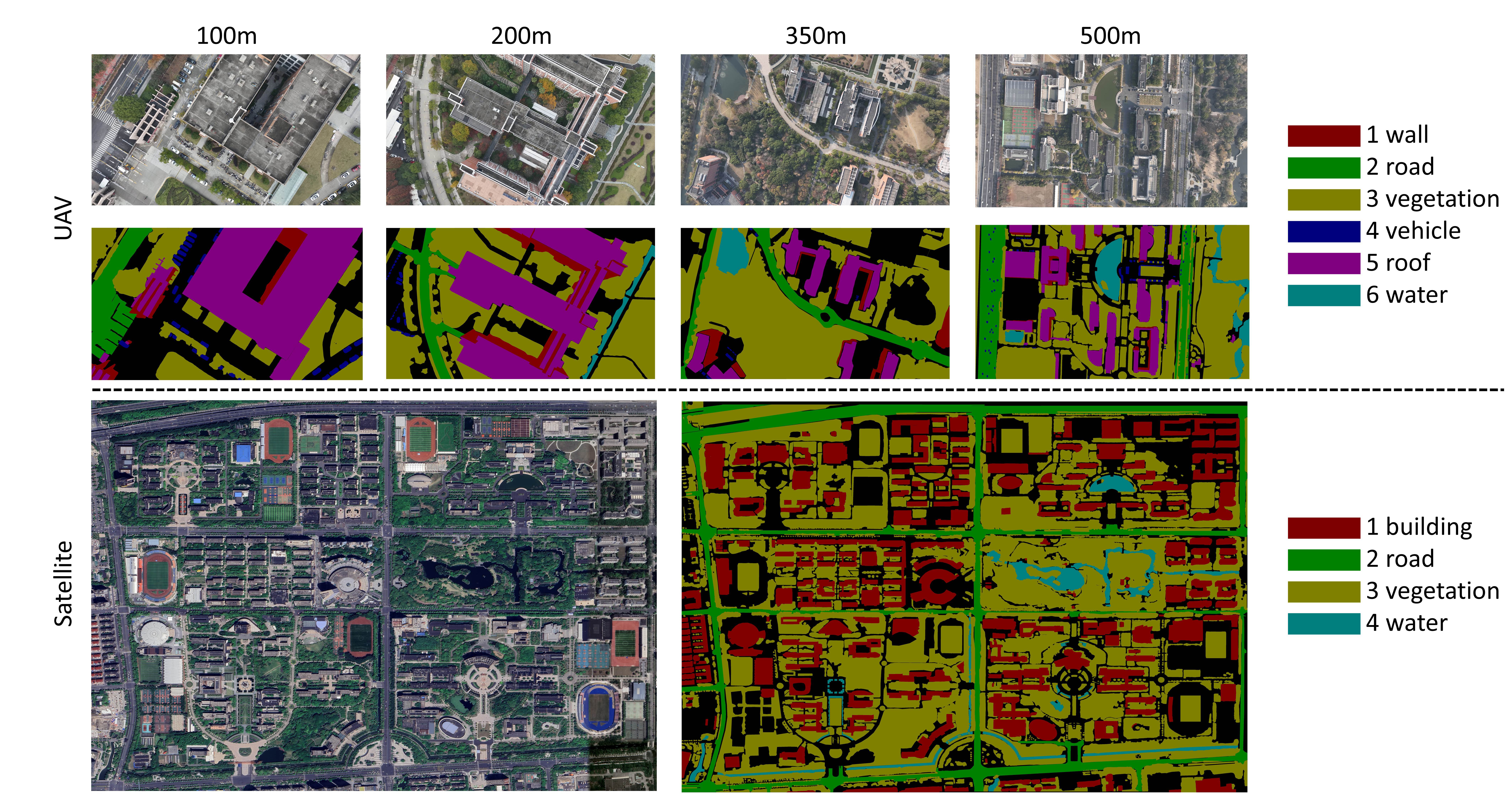}
    \caption{Examples from MAFS and SemanticMAFS. The top row shows UAV images annotated into six categories: walls, roads, vegetation, vehicles, roofs, and water. The bottom row shows satellite images annotated into four categories: buildings, roads, vegetation, and water.}
    \label{fig3}
\end{figure*}

The dataset covers aerial imagery from 14 university campuses in Hangzhou (MAFS-00 to MAFS-13) and one large-scale 2 km × 2 km area (MAFS-14). As shown in Fig. \hyperref[fig2]{2}, two flight strategies are employed: fixed-altitude and variable-altitude. UAV flights span 100 to 500 meters, capturing geographic and environmental variations across altitudes and temporal conditions.  

Aerial videos are recorded with a DJI MAVIC III E, synchronized with onboard IMU sensor data. Each frame includes geospatial metadata such as latitude, longitude, altitude, yaw, and velocity. All footage is captured in 4K resolution at 30 fps, ensuring high visual fidelity and ecological detail. The surveyed areas feature diverse landmarks, including buildings, roads, vegetation, and water bodies, providing a challenging and representative dataset. For each flight path, corresponding Google Maps Level 19 imagery is provided as a geolocation reference. Experimental validation confirms the robustness of this dataset in supporting UAV localization research.

\subsection{UAV Top-view Image Semantic Dataset: SemanticMAFS}

The SemanticMAFS dataset extends MAFS by incorporating semantic annotations. It consists of two components:  

1) UAV top-view images sampled from MAFS at multiple altitudes, annotated into seven semantic categories: roof, wall, ground, road, vegetation, water body, and vehicle.  
2) Satellite imagery corresponding to MAFS flight paths, annotated into four categories: building, ground, road, and vegetation.  

Annotation is performed using the Eiseg interactive tool \cite{bib36}. SemanticMAFS poses unique challenges: as UAV altitude increases, the field of view expands while ground-level details remain discernible, making segmentation more complex. On average, annotating a single image requires over 60 minutes. Representative examples are shown in Fig. \hyperref[fig3]{3}, illustrating both UAV and satellite annotations.

\section{Method}

\begin{figure*}[!b]
    \centering
    \includegraphics[width=0.9\textwidth]{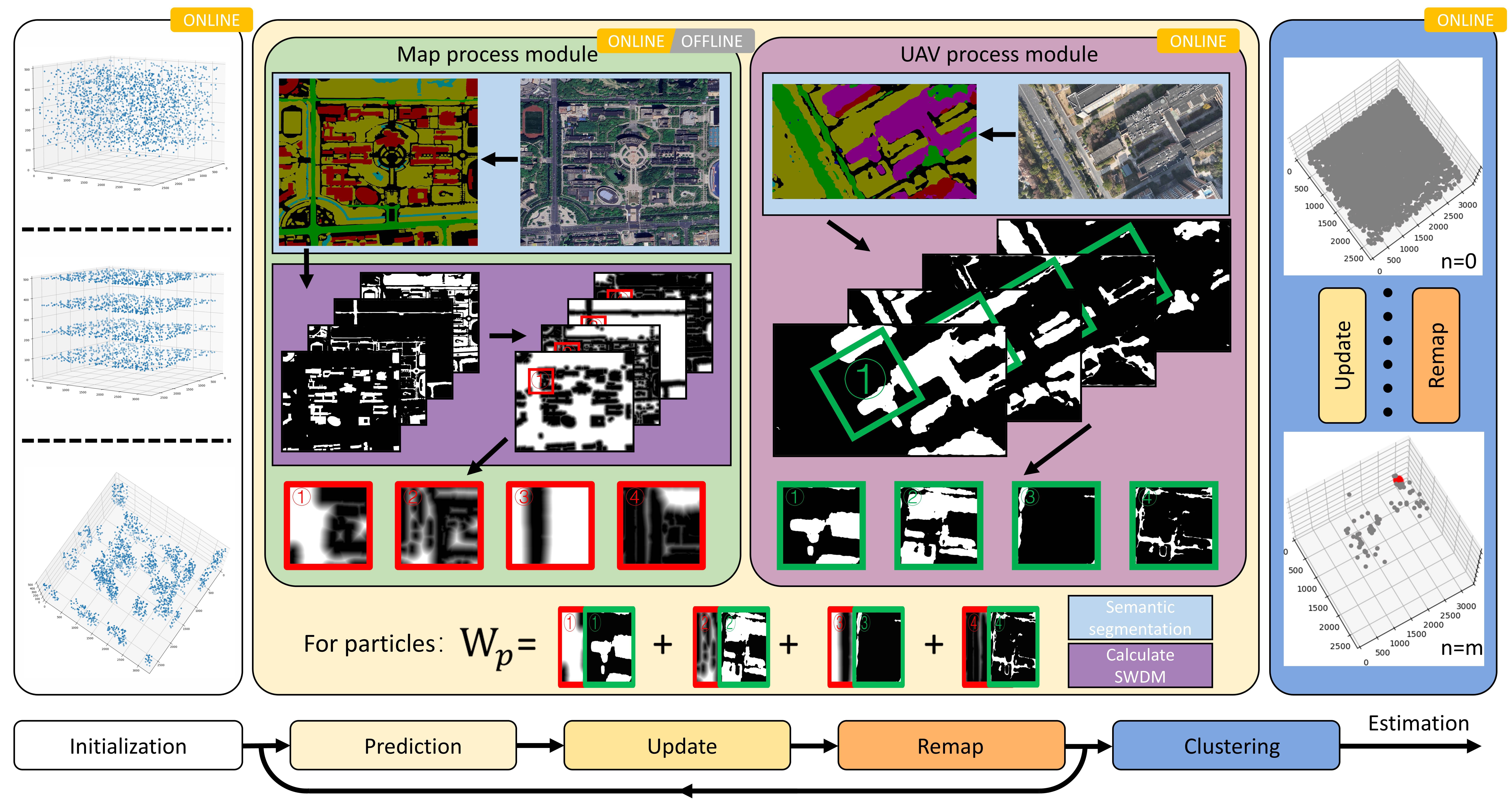}
    \caption{Overall framework of the proposed SWA-PF. Particle states are initialized using prior knowledge. During prediction, real-time semantic segmentation of UAV top-view images is performed to compute particle weights. Through iterative updating and mapping, UAV position is estimated via particle clustering.}
    \label{fig4}
\end{figure*}

Inspired by the established success of Monte Carlo Localization (MCL) in robotic positioning \cite{Mc1}, this study adopts a Particle Filter (PF) as the core framework for visual-inertial fusion localization in UAV systems. The particle filter is particularly adept at handling multi-modal state distributions, iteratively updating a set of particles to represent potential system states. As the UAV navigates its environment and gathers sensor data, these particles continuously refine the platform's pose estimation via Bayesian updating. However, while this probabilistic approach is widely used in robotic localization, its computational demands present implementation challenges. To mitigate this, we introduce two key components: a semantic segmentation module for UAVs operating at varying altitudes and a semantic-weighted adaptive particle filter localizer, as shown in Fig. \hyperref[fig4]{4}. This section details the objectives and implementation of these components.

\subsection{Semantic Segmentation}

\subsubsection{Semantic Segmentation of Satellite Images}
A VGG-pretrained \cite{bib37} U-Net \cite{bib38} architecture is employed for satellite image segmentation. The network is trained on three annotated satellite images from the SemanticMAFS dataset. During preprocessing, the images, resized to 256×256px, are fed into the segmentation network to standardize input dimensions. Data augmentation techniques such as random scaling, rotation, cropping, and flipping are applied to enhance dataset diversity and size. Specifically, random scaling improves generalization for images captured at varying altitudes. Despite being trained on only four images, the model demonstrates satisfactory segmentation performance, which can be attributed to the dense object instances within each image and the shared urban context between the training and test datasets, reducing the need for extensive generalization. The model's performance metrics and segmentation results are presented in the upper part of Fig. \hyperref[fig5]{5}.

\subsubsection{Semantic Segmentation of UAV Images}
Segmenting UAV top-view images introduces more significant challenges than satellite image segmentation due to the variable altitudes of UAVs, often leading to incomplete representations of buildings. Initially, a U-Net architecture was used to classify features into four categories: buildings, ground, roads, and vegetation. However, preliminary tests revealed critical misclassifications, such as rooftops being incorrectly identified as ground surfaces, undermining localization accuracy. To address this, we adapt the methodology from the VDD dataset \cite{VDD}, enhancing the SemanticMAFS dataset by distinguishing buildings into walls and roofs and adding two more classes (water bodies and vehicles) for network training.

\begin{figure*}[!t]
    \centering
    \includegraphics[width=.9\textwidth]{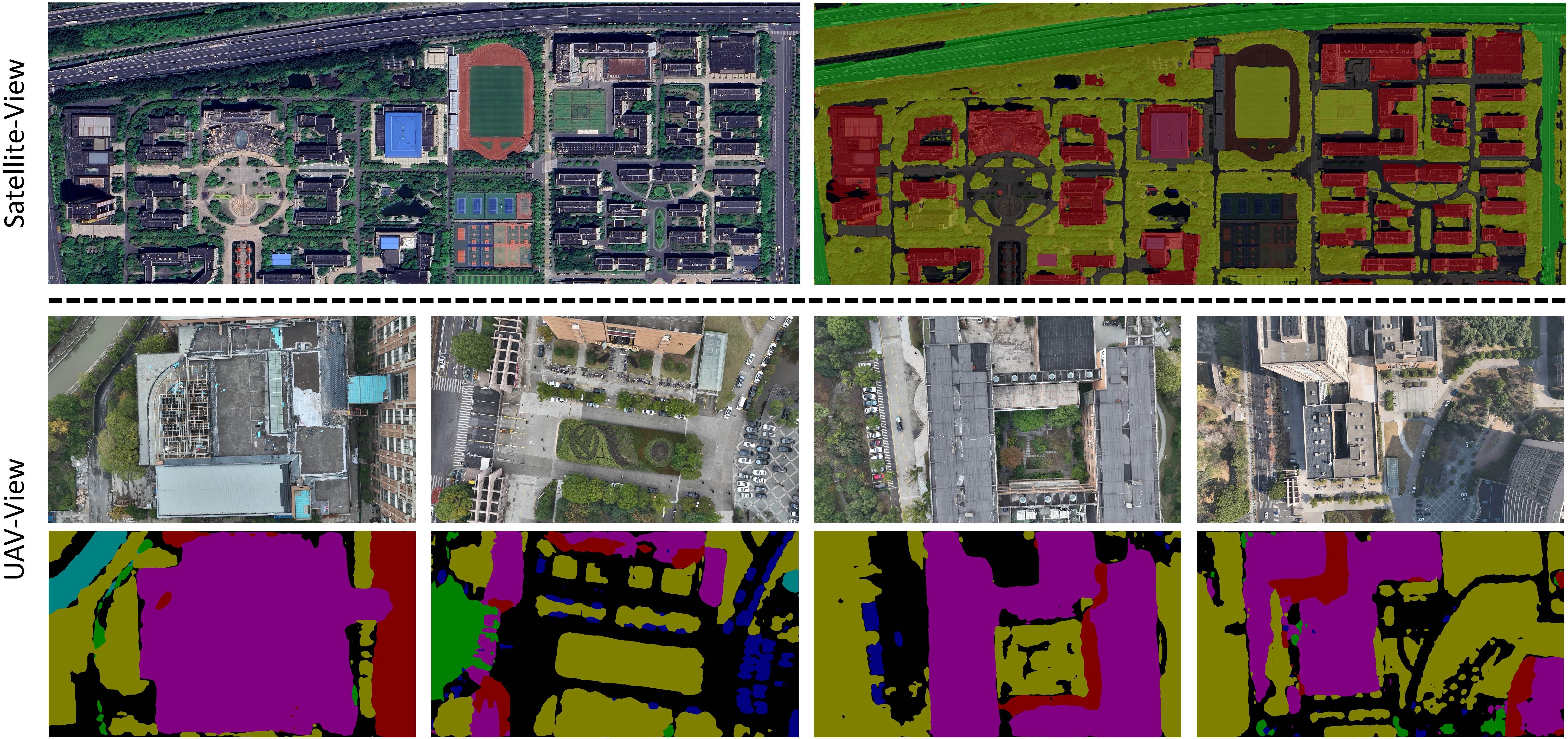}
    \caption{Results of semantic segmentation on satellite and UAV images using U-Net \cite{bib38} and SegFormer \cite{bib40}, which display substantial realistic noise.}
    \label{fig5}
\end{figure*}

After reviewing recent advances in semantic segmentation, we selected the SegFormer-B0\cite{bib40} architecture due to its optimal trade-off between accuracy and processing speed. The model was developed using PyTorch and trained on a single NVIDIA GeForce RTX 4060Ti GPU (16 GB VRAM) in an environment with Python 3.8, PyTorch 1.10.1, and CUDA 11.3. The network input size was set to 512×512 pixels, and the model was initialized with weights pre-trained on ImageNet. Training occurred in two stages: first, the backbone network was frozen and trained for 50 epochs with a batch size of 32. Then, the entire network was unfrozen and fine-tuned for an additional 200 epochs with a batch size of 8. We utilized the AdamW optimizer (initial learning rate = 1e-4, weight decay = 1e-2) with a cosine annealing learning rate schedule. This setup resulted in a final mean intersection-over-union (mIoU) of 78.80\% and an overall pixel accuracy of 91.41\%.

The lower part of Fig. \hyperref[fig5]{5} showcases the output of our trained network. Although these results do not reach state-of-the-art levels, robust semantic segmentation is not the focus of this study. Notably, both UAV top-view and satellite images exhibit significant real-world noise.

\subsection{Semantic-Weighted Adaptive Particle Filter}

\subsubsection{Problem Formulation}

Since satellite images are two-dimensional, estimating the 4DOF (latitude, longitude, altitude, and yaw) for drone localization requires extending the 2D problem into 3D space. Thus, we expand the particle distribution from a 2D plane into 3D and define the state space based on the drone's pose within this volume. The state vector is expressed as \( x_{t} = \begin{bmatrix} x & y & h & \theta \end{bmatrix} \), where \( x \), \( y \), and \( h \) represent the 3D positional coordinates, and \( \theta \) denotes the yaw angle.

\subsubsection{Prediction}

The system incorporates a control input, \( u_t \), representing the rigid-body transformation derived from odometry data since the previous frame. This input includes linear velocity \( v_t \) and angular velocity \( \omega_t \). Simultaneously, at each time step \( t \), an observational measurement \( z_t \) is obtained by processing semantic scans from the drone's downward-facing camera.

Based on Bayesian theory, the filter recursively updates the posterior probability distribution of the state, fusing information from the system dynamics model and incoming observations. The core algorithm consists of two sequential steps:

\begin{equation} 
\small
P(x_t|z_{1:t-1}, u_{1:t}) = \int P(x_t|x_{t-1}, u_t) P(x_{t-1}|z_{1:t-1}, u_{1:t-1}){\rm d} x,
\end{equation}

\begin{equation} 
P(x_t|z_{1:t-1}, u_{1:t}) = \frac{P(z_t|x_t) P(x_t|z_{1:t-1}, u_{1:t})}{P(z_t|z_{1:t-1}, u_{1:t})}.
\end{equation}

The first phase utilizes a state transition model to predict the prior distribution at the current time step based on the posterior distribution \( p(x_{t-1} | z_{1:t-1}, u_{1:t-1}) \) from the preceding step. Upon receiving a new observation \( z_t \), the prior distribution is updated using Bayes' theorem by integrating the observation likelihood to compute the posterior distribution.

The state of the particles is then updated according to the kinematic model:

\begin{align} 
    P(x_t|x_{t-1}, u_t) &= x_{t-1} + v_{t-1}\Delta t + \omega_{t-1}\Delta t + \epsilon,
\end{align}

where the variables \( x \), \( y \), \( h \), and \( \theta \) are corrupted by an additive noise vector \( \epsilon \), which is governed by a Gaussian distribution \( \epsilon \sim \mathcal{N} (0, \sigma^2) \).

\subsubsection{Measurement Model}

In our measurement model, the observation \( z_t \) is a semantic UAV top-view image, represented by the triplet \( \{P_i, P_j, L_{ij}\} \), where \( P_x \) and \( P_y \) denote the image coordinates and \( l_0 \) represents the semantic label. For each particle state, we query the semantic map \( M \) to extract the expected semantic category.

\begin{figure}[!h]
    \centering
    \includegraphics[width=.95\linewidth]{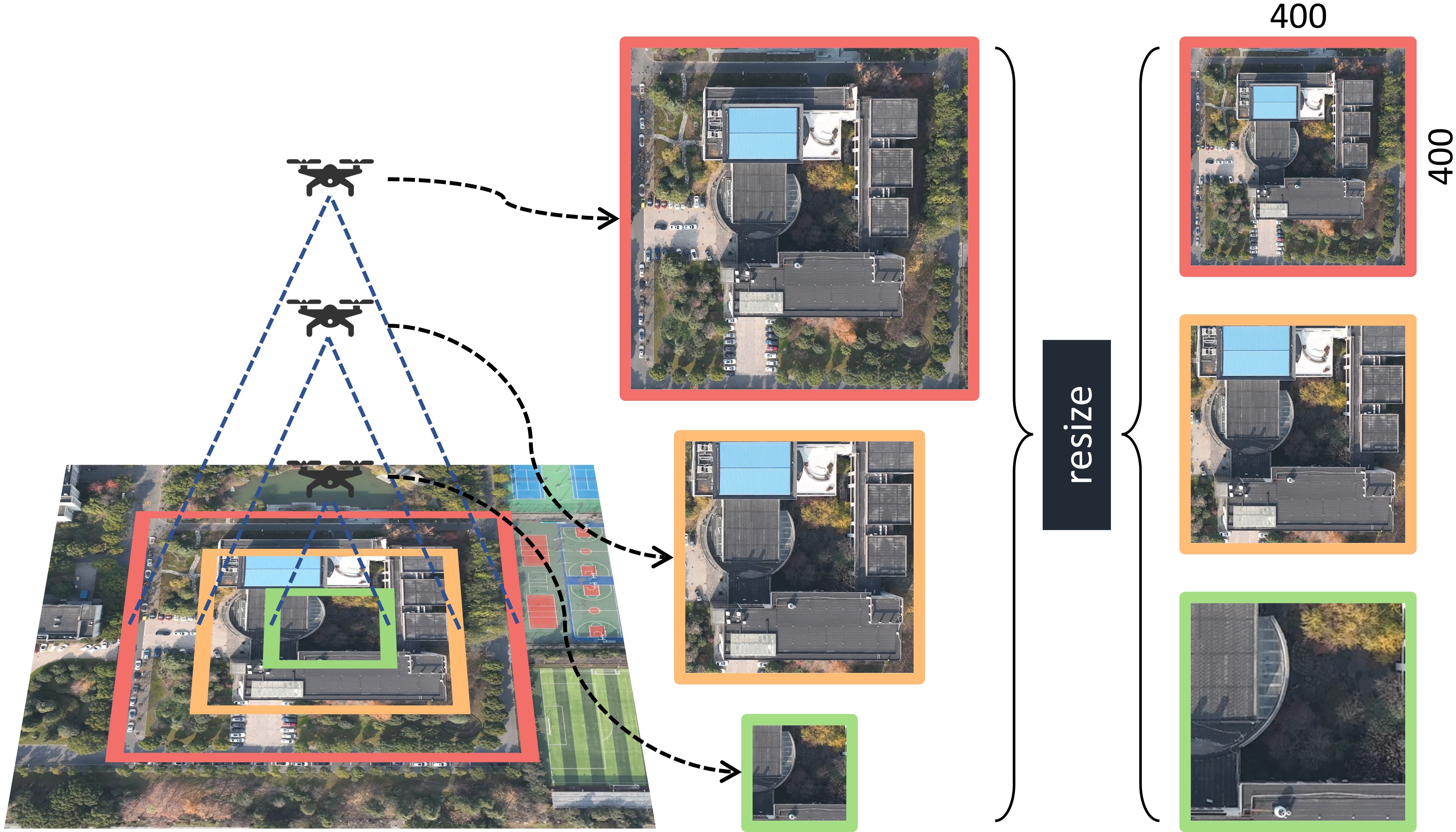}
    \caption{Particles at different altitudes exhibit distinct fields of view, which can significantly affect particle weight calculations. Our approach resizes all particle viewpoints to a fixed size of 400 × 400.}
    \label{fig6}
\end{figure}

To mitigate significant variations in the field of view caused by changes in altitude \( h \) at the same (\( x \), \( y \)) coordinates (as shown in Fig. \hyperref[fig6]{6}), we scale both the UAV-view image \( z_t \) and the cropped semantic map \( M \) to a fixed size of 400 × 400 pixels.

\vspace{-10pt}

Fig. \hyperref[fig7]{7} illustrates the Semantic Weighted Distance Map (SWDM) for the MAFS-10 path within satellite imagery. Instead of calculating semantic consistency rates between particle and UAV perspectives, we utilize the Semantic-weighted Distance Map. The semantic-weighted distance map \( M_{(nearest|p)} \) is computed as:

\begin{figure*}[!t]
    \centering
    \includegraphics[width=\textwidth]{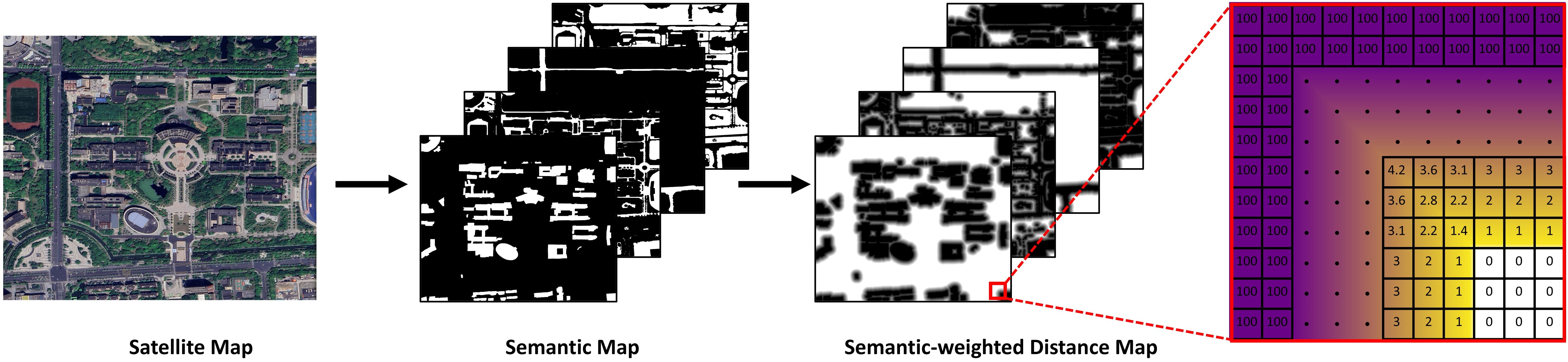}
    \caption{The classification semantic map and the Semantic Weighted Distance Map (SWDM). After semantic segmentation of satellite imagery, the SWDM is calculated for each layer. This computation can be executed either during preprocessing or in real-time processing systems.}
    \label{fig7}
\end{figure*}

\begin{equation} 
    M_{(nearest|p)} = \sum_{L}\sum_{i=0}^{H} \sum_{j=0}^{W} \min_{L_{ij}=L} ||p_{ij}-p||_2.
\end{equation}

This metric calculates the Euclidean distance to the nearest semantically equivalent point across all pixels in the image.

\begin{figure}[!h]
    \centering
    \includegraphics[width=.85\linewidth]{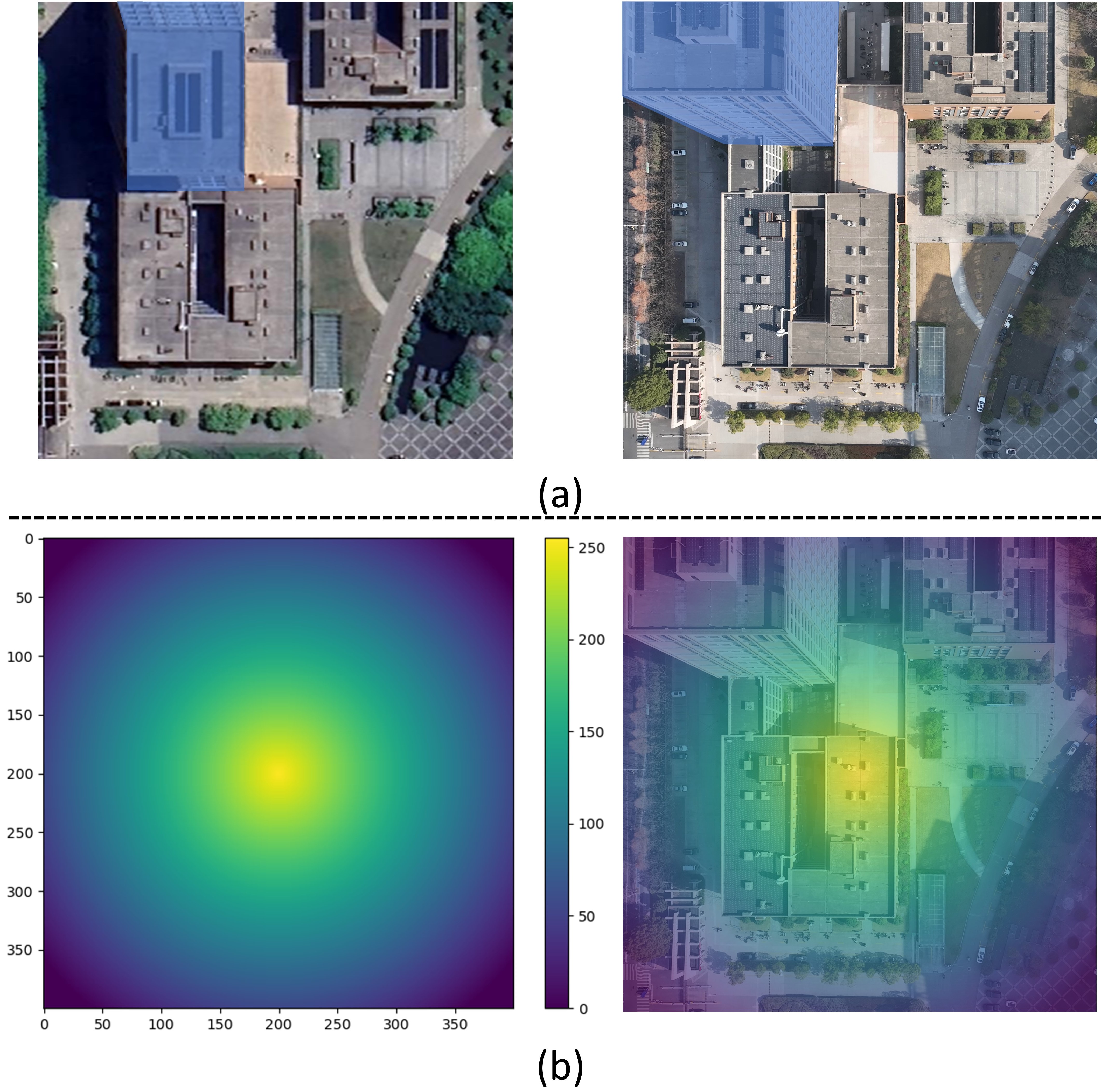}
    \caption{(a) An example demonstrates that even the same building exhibits significant differences in style and perspective between UAV and satellite image. (b) Center distance field matrix \(M_{CDF}\). We use \(M_{CDF}\) to increase the center weight and encourage center alignment.}
    \label{fig8}
\end{figure}

To enhance vertical convergence and emphasize central semantic similarities, we incorporate a center distance field matrix \( M_{CDF} \) (as shown in Fig. \hyperref[fig8]{8}). The weight \( W \) for each particle is then computed as:

\begin{equation} 
    W = \sum_{p} \frac{\alpha_p}{z_t \cdot M_{(nearest|p)}} \cdot M_{CDF} + \gamma.
\end{equation}

Here, \( \alpha_p \) is a weight parameter specific to each semantic class, and \( \gamma \) is a normalization constant that slows convergence. Finally, all particle weights are normalized such that their sum equals 1.

\vspace{-10pt}

\subsection{Performance Optimization}

Initial performance limitations arise from the computational demands of image rotation and translation. Each particle requires individual pose adjustments involving both translation and rotation. This process is especially resource-intensive when handling large particle quantities due to the computational complexity of rotation matrices, and may also lead to information degradation during rotational transformations.

\begin{figure}[!b]
    \centering
    \includegraphics[width=.95\linewidth]{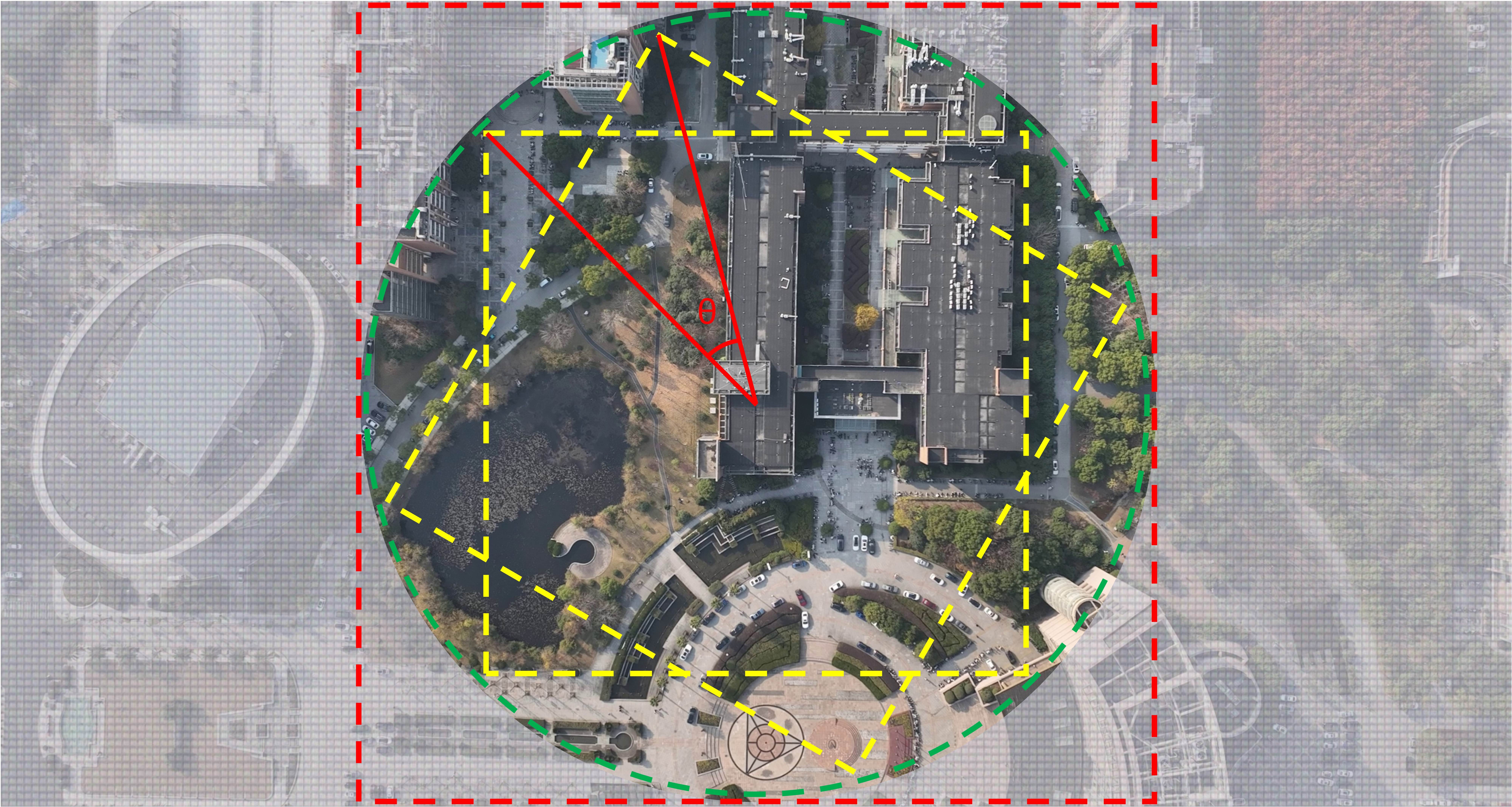}
    \caption{We match the direction angle of its particles with the nearest angle. The yellow square represents the saved rotated image, and the gray area represents the discarded part. \( \theta \)’s value is 3.6.}
    \label{fig9}
\end{figure}

Our strategy divides the UAV-view image rotation into 100 discrete steps at 3.6° intervals. To optimize computational time, particles are weighted based on yaw angle matching with the nearest rotational perspective. During rotation, spatial constraints require selective exclusion of image regions. This process is illustrated in Fig. \hyperref[fig9]{9}. Although this approach discards peripheral image data, the matrix \( M_{CDF} \) weighting method minimizes information loss by progressively reducing label weights towards the image boundaries.

Further optimizations include matrix operation simplification using two-stage matrix multiplication and accelerated batch processing through optimized matrix stacking. The GPU-accelerated implementation processes 20,000 particle matrices in approximately 4.9 seconds. Detailed performance benchmarks will be presented in the subsequent section.

When estimating UAV positions using particle swarm fitting, directly calculating the mean of posterior particle filters may introduce significant localization errors due to outliers. To address this, we apply outlier detection filtering. Common techniques include distance-based methods, density-based approaches (e.g., DBSCAN \cite{bib50}), and statistical analyses (e.g., Z-score). Given the potential clustering of particles, we select DBSCAN for its ability to identify high-density clusters while flagging low-density points as outliers. The implementation follows four steps: First, DBSCAN removes outliers. Next, the remaining particles are clustered. Then, the cluster centroids are computed. Finally, we monitor the posterior particle filter's covariance. When it falls below a predefined threshold, we define the cluster centroid as the position estimate and locate it on the global map.


\begin{table*}[!b]
    \caption{The running time of different methods on MAFS-10}
    \label{tab2}
    \resizebox{1.0\textwidth}{!}{
    \begin{tabular}{lccccccccccc}
    \hline
     & \textbf{SIFT}\cite{SIFT} & MI\cite{MI} & HIST\cite{histogram} & AKAZE\cite{AKAZE} & BRISK\cite{BRISK} & ORB \cite{ORB} & SSPT\cite{SSPT} & DenseUAV\cite{DenseUAV} & OS-FPI\cite{OSFPI} &\textbf{SWA-PF(ours)}\\ \hline
    Fitting time (s)  & 1630    & 433     & 143     & 371     & 1191    & 90     & -       & -        & -      & {\textbf{7}} \\ 
    Finish time (s)   & 3338    & 1301    & 524     & 858     & 3028    & 299    & -       & -        & -       & {\textbf{25}}\\ 
    RMSE (m)            & 62.825 & 21.775 & 14.317 & 16.401 & 16.469 & 21.11 & 65.958 & 359.416 & 90.114  & {\textbf{6.5685}}\\ 
    Recall@10 (\%)     & 0       & 5.263  & 9.210  & 34.210 & 6.578  & 7.894 & 34.210 & 1.315   & 14.473 & {\textbf{97.368}}\\ 
    Median (m)         & 38.82   & 14.82   & 12.12   & 11.77   & 16.02   & 17.57  & 14.04   & 296.1    & 87.43   & {\textbf{6.653}}\\ 
    \hline
    \end{tabular}}
\end{table*}

\begin{figure*}[!b]
    \centering
    \begin{minipage}[t]{\textwidth}
        \centering
        \includegraphics[width=0.9\linewidth]{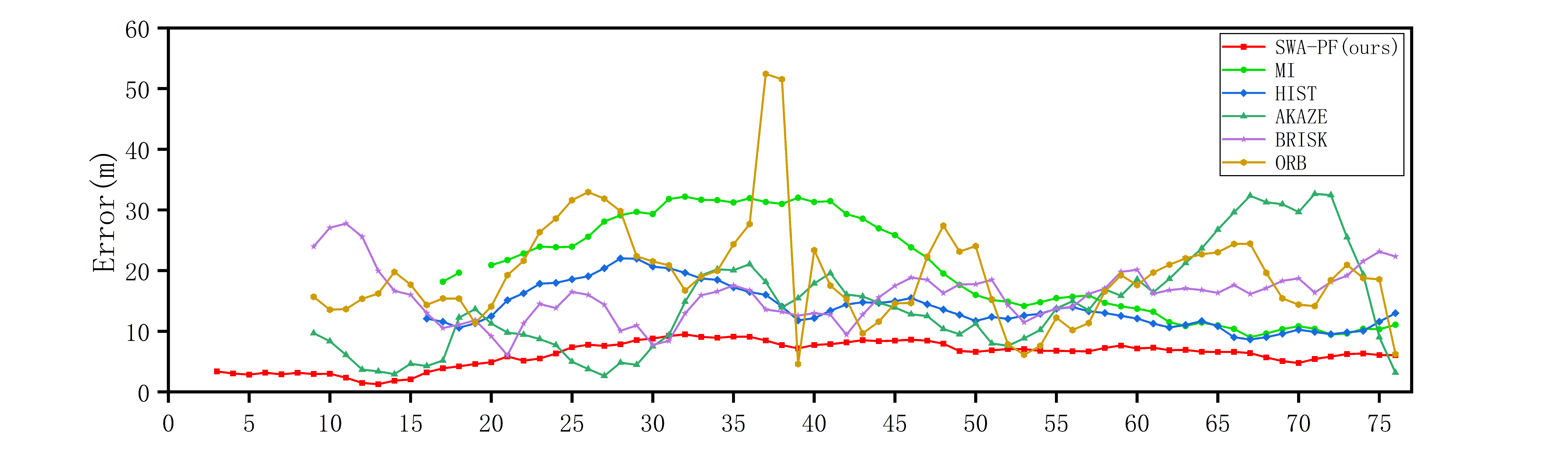}
    \end{minipage}
    \begin{minipage}[t]{\textwidth}
        \centering
        \includegraphics[width=0.9\linewidth]{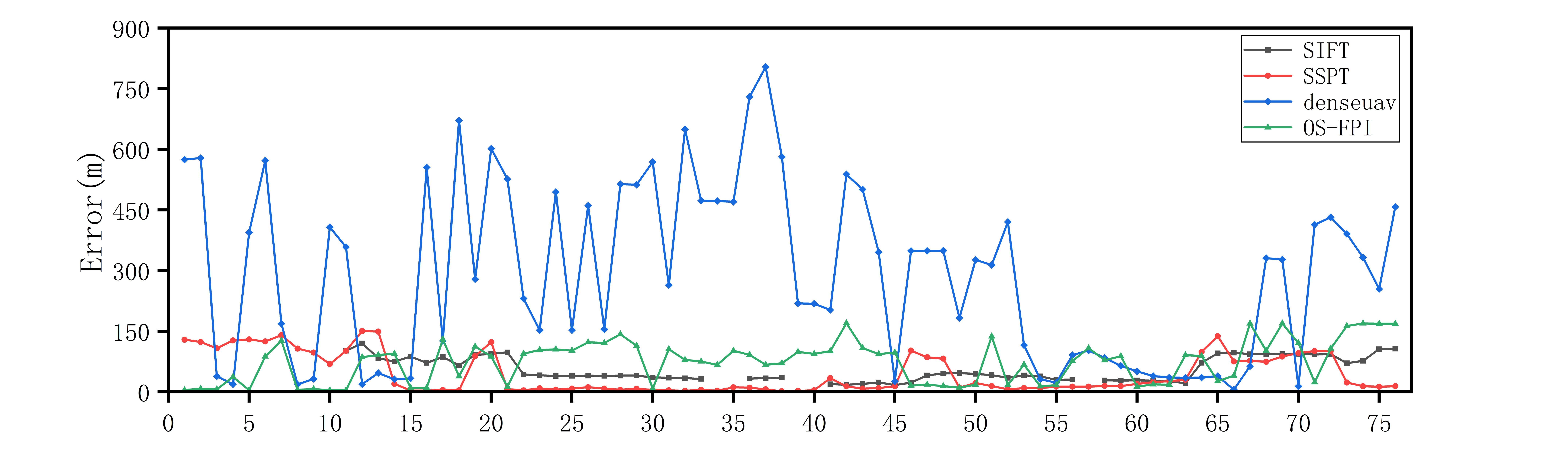}
    \end{minipage}
    \caption{The lowest global location error achieved by each evaluated method on the MAFS-10 dataset at a fixed altitude of 200 meters.}
    \label{fig10}
\end{figure*}

\section{Experiment}
\label{section IV}
This section presents a comprehensive evaluation of the proposed SWA-PF method. First, we benchmark SWA-PF against conventional particle filter-based localization algorithms (Section \hyperref[section IVA]{IVA}). Second, ablation studies validate critical design components (Section \hyperref[section IVB]{IVB}). Finally, a comparative analysis with state-of-the-art methods demonstrates the performance superiority of our approach (Section \hyperref[section IVC]{IVC}).

\subsection{Comparison of Particle Filter-based Localization Algorithms}
\label{section IVA}

\subsubsection{Verification Setup}
We evaluate our method using three distinct trajectories from the MAFS dataset: two standardized flight scenarios (MAFS-03 and MAFS-10) characterized by fixed-altitude, constant-velocity patterns at 200 meters, and one complex dynamically challenging scenario (MAFS-14) involving a variable altitude trajectory ranging from 150 meters to 500 meters. To ensure consistency in data processing across methods, still images are extracted from the original video stream at a fixed frequency. Each trajectory undergoes 20 experimental trials, during which both optimal and average performance metrics are documented. For MAFS-03 and MAFS-10, particle filter initialization uses 5,000 particles, while the larger spatial search domain of MAFS-14 requires 40,000 particles.

To systematically evaluate traditional feature extraction algorithms in UAV visual localization, we compare six representative methods: Mutual Information (MI) template matching \cite{MI}, HIST-based matching \cite{histogram}, Scale-Invariant Feature Transform (SIFT) \cite{SIFT}, binary-optimized AKAZE \cite{AKAZE}, Binary Robust Invariant Scalable Keypoints (BRISK) \cite{BRISK}, and Oriented FAST and Rotated BRIEF (ORB) \cite{ORB}. Results are compared with our proposed SWA-PF framework.

For the template-based approach, the Normalized Cross-Correlation (NCC) \cite{bib44} is computed between poses to assess similarity. For HIST-based methods, grayscale histograms of paired images are constructed and compared via NCC. Regarding feature-based methods, the workflow involves: (1) feature extraction from image pairs, (2) feature matching using the K-Nearest Neighbor (KNN) method, and (3) homography estimation with inlier counting. The inlier count serves as the similarity metric, where higher values indicate greater correspondence. Given the rotational invariance of these features, we constrain particle orientation angles by assuming a fixed UAV heading.

Additionally, we evaluate three more methods: OS-FPI \cite{OSFPI}, DenseUAV \cite{DenseUAV}, and SSPT \cite{SSPT}. DenseUAV is an image retrieval technique; we tile the satellite map into UAV-sized patches with 50\% overlap and select the best-matching patch's center as the estimated position. OS-FPI and SSPT are end-to-end deep learning methods. Following the authors' recommendations, we crop a 3x UAV-sized neighborhood around the true location and report the network's predicted center as the localization error.

Particle noise parameters for motion updates are set to \( \epsilon = 15 \), intentionally exceeding quadcopter IMU error thresholds to create a rigorous validation environment. A uniform parameter \( \gamma = 10 \) is maintained across all experimental runs.

\begin{figure*}[!b]
    \centering
    \includegraphics[width=0.9\textwidth]{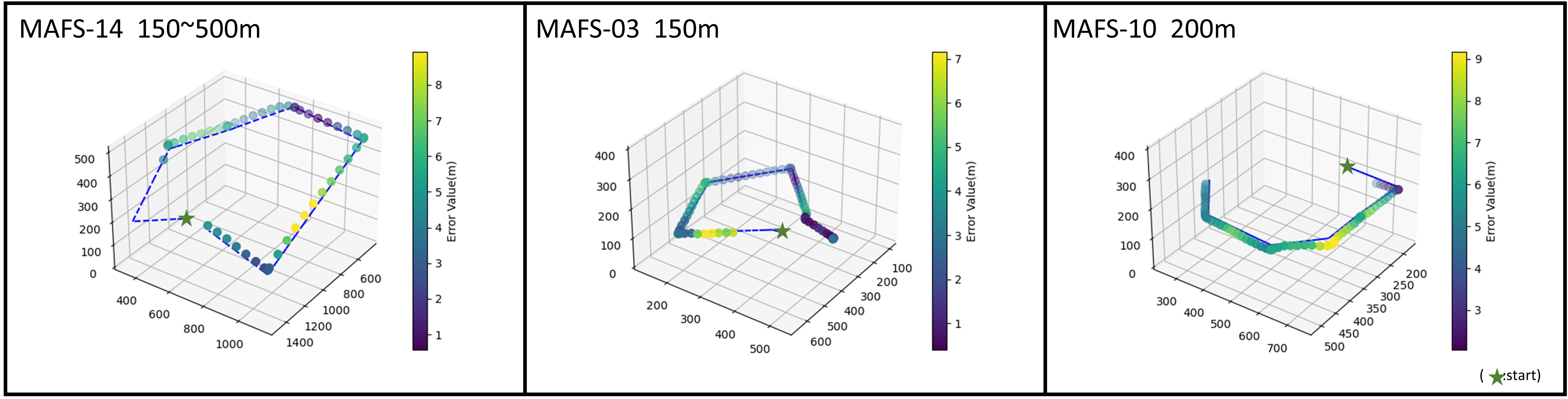}
    \caption{3D path diagrams of SWA-PF in MAFS-03, MAFS-10, and MAFS-14. The variable-altitude configuration is used for MAFS-14, and fixed-altitude configurations for MAFS-03 and MAFS-10. The color bars represent the error value in meters, with point colors corresponding to the error value. The red star indicates the starting point of each trajectory.}
    \label{fig11}
\end{figure*}

\subsubsection{Computational Efficiency}
We evaluate inference times for various image matching methods using the MAFS-10 benchmark. All implementations use multithreaded processing for performance enhancement, with results summarized in Table \hyperref[tab2]{II}. Despite the computational complexity inherent in semantic feature extraction, our method shows a significant speed advantage, thanks to optimized GPU acceleration combined with multithreaded architecture. Notably, all experiments—including map processing, semantic segmentation, and localization—are conducted in real-time using an i5-12400F CPU and NVIDIA GeForce RTX4060Ti GPU configuration, demonstrating deployability on medium-sized UAVs.

\subsubsection{Localization Accuracy}
Fig. \hyperref[fig10]{10} compares the minimal global localization error of the evaluated methods on the fixed-altitude MAFS-10 dataset. Several comparative methods yield ambiguous pose estimations, whereas our method consistently selects the optimal solution as the final output. Among traditional methods, BRISK and HIST show higher localization accuracy; however, their high computational expense prevents them from meeting real-time processing requirements. Conversely, the ORB method operates in real-time but suffers from lower positioning accuracy and slower convergence. Methods such as SSPT, DenseUAV, and OS-FPI, which process each frame independently, often exhibit discontinuous estimation errors and temporal instability when applied to continuous video, a significant limitation for practical real-world applications. Our semantic-based measurement model, in contrast, achieves more accurate estimations with improved computational efficiency, converging in 7 seconds and processing the entire trajectory in 30 seconds.

\subsubsection{Robustness Under Variable Altitude}
In the MAFS-14 variable-altitude scenarios, particle convergence proves significantly more challenging. Conventional methods exhibited near-complete failure, aligning with theoretical expectations. In contrast, our extended three-dimensional particle filter implementation maintains operational efficiency despite the height variation challenges. Although three-dimensional localization slightly affects convergence rates, the required iteration count remains comparable to fixed-altitude scenarios. This performance preservation is attributed to our optimized weighting function design, which exhibits exceptional sensitivity to the vertical dimension while maintaining computational efficiency. Fig. \hyperref[fig11]{11} presents the three-dimensional path diagrams of our results in MAFS-03, MAFS-10, and MAFS-14. Our method shows excellent convergence in the vertical direction. Overall, our approach provides superior UAV positioning accuracy compared to other methods.

\subsection{Ablation Study}
\label{section IVB}

\subsubsection{Positioning Error Analysis}

\begin{figure}[!h]
    \centering
    \includegraphics[width=0.9\linewidth]{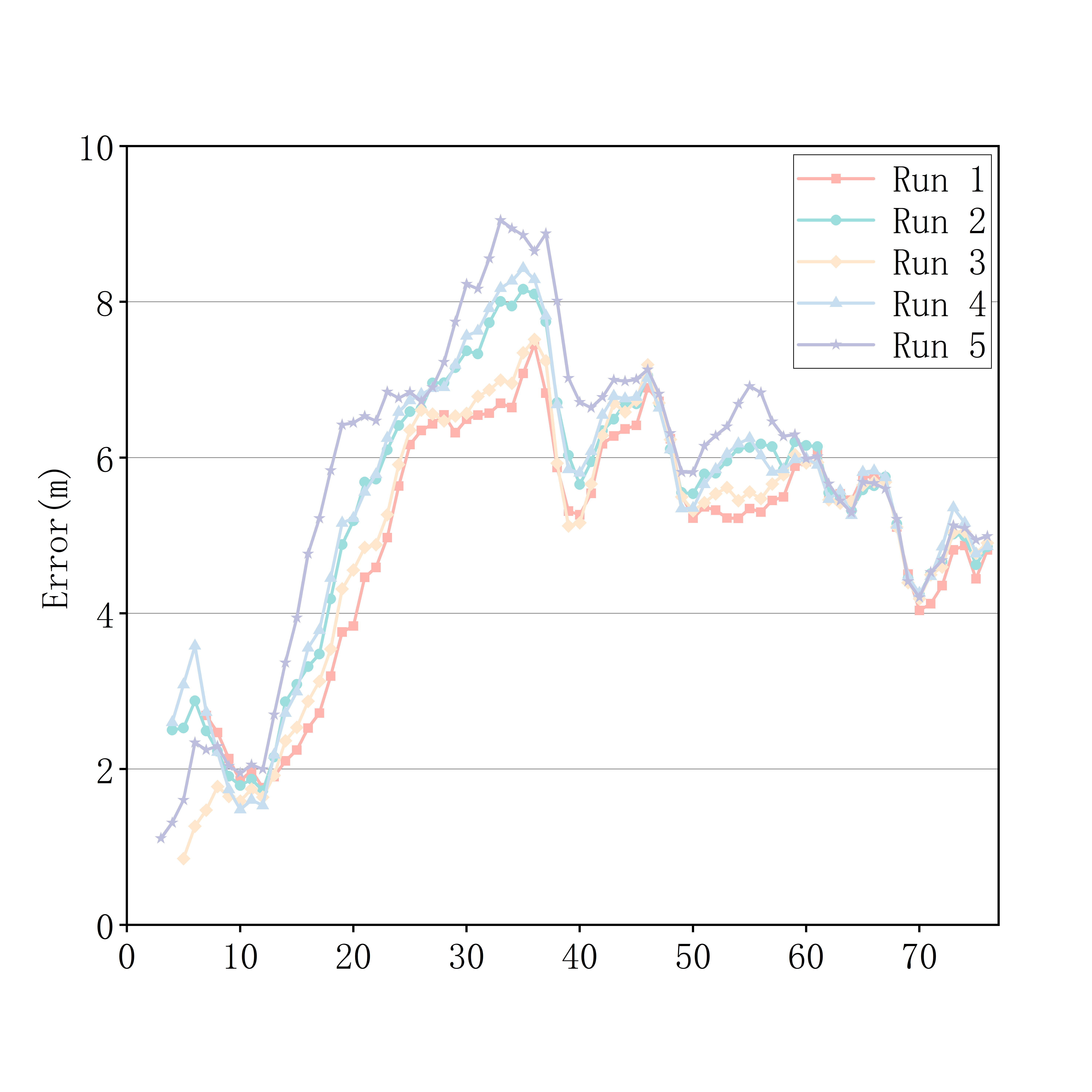}
    \caption{The lines 1-5 represent the error results from five experiments using the SWA-PF method on the same trajectory. The overall errors exhibit a consistent trend.}
    \label{fig12}
\end{figure}

Our systematic analysis of localization errors (Fig. \hyperref[fig12]{12}) reveals consistent spatial error patterns across repeated path trials. The data indicates significant mid-trajectory positioning inaccuracies, with errors clustered in specific spatial regions. Two primary factors contribute to these observations: (1) Perspective disparity between oblique satellite imagery and UAV top-view images, leading to substantial geometric mismatches that amplify positioning errors. (2) Limited distinctive visual features along the test path, reducing the availability of reference points for spatial alignment. While our method is unable to accurately estimate pose in feature-scarce scenarios (e.g., when large portions of the field of view are dominated by buildings or vegetation), this limitation is common to many existing visual localization techniques that struggle in such environments.

\begin{figure}[!h]
    \centering
    \includegraphics[width=0.9\linewidth]{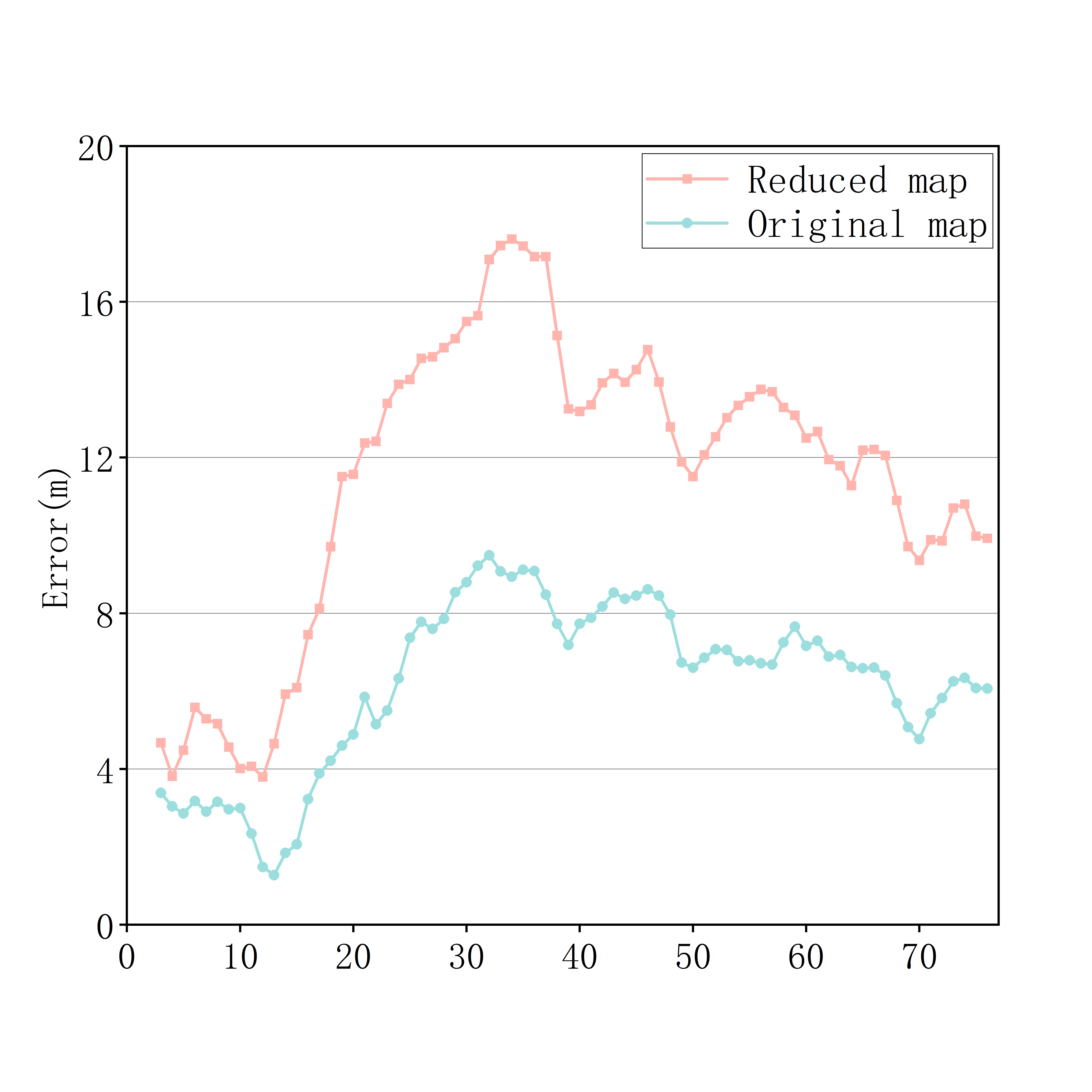}
    \vspace{-10pt}
    \caption{Comparison of SWA-PF method error in original and low-resolution images. The pixel size of the map was reduced by half, leading to a doubling of the pixel error.}
    \label{fig15}
\end{figure}



\subsubsection{Performance Under Low Resolution and Temporal Map Variations}

Our methods, based on semantic satellite maps, exhibit strong resistance to interference when significant temporal variations occur between satellite images. Additionally, our approach performs well in low-resolution environments. We tested our system with satellite images at half the original resolution, and the resulting error showed minimal change. These results are depicted in Fig. \hyperref[fig15]{13}.

\subsubsection{Impact of Particle Initialization Methods on Convergence Rate}

\begin{figure}[!h]
    \centering
    \includegraphics[width=.95\linewidth]{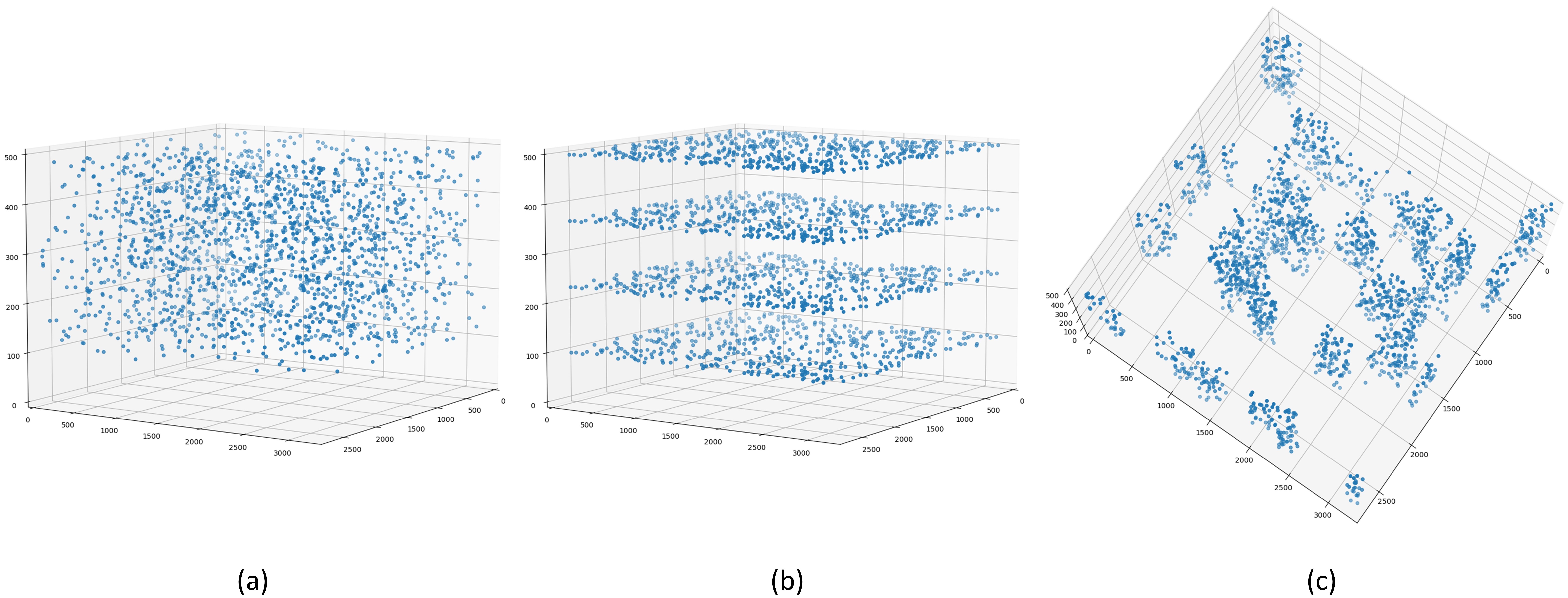}
    \caption{Three different particle initialization methods: (a) full-space randomization, (b) layered randomization, (c) center semantic class alignment.}
    \label{fig13}
\end{figure}

We implement three distinct particle filter initialization methods (Fig. \hyperref[fig13]{14}) leveraging semantic understanding and prior geographic knowledge. Method (a) employs full-space randomization, dispersing particles randomly across the 3D environment. Method (b) uses layered randomization, where particles are distributed within a single layer and duplicated across multiple layers. Method (c) aligns particle clusters with semantic classes concentrated in the UAV's central visual field. Our tests reveal that method (c) achieves faster initial convergence when UAV imagery contains clear semantic features. Although method (b) reduces initialization time, significant discrepancies between preset layer heights and actual UAV altitude compromise its convergence performance.

\subsubsection{Optimization Performance of Adding Center Distance Field}

\begin{figure}[!h]
    \centering
    \includegraphics[width=.95\linewidth]{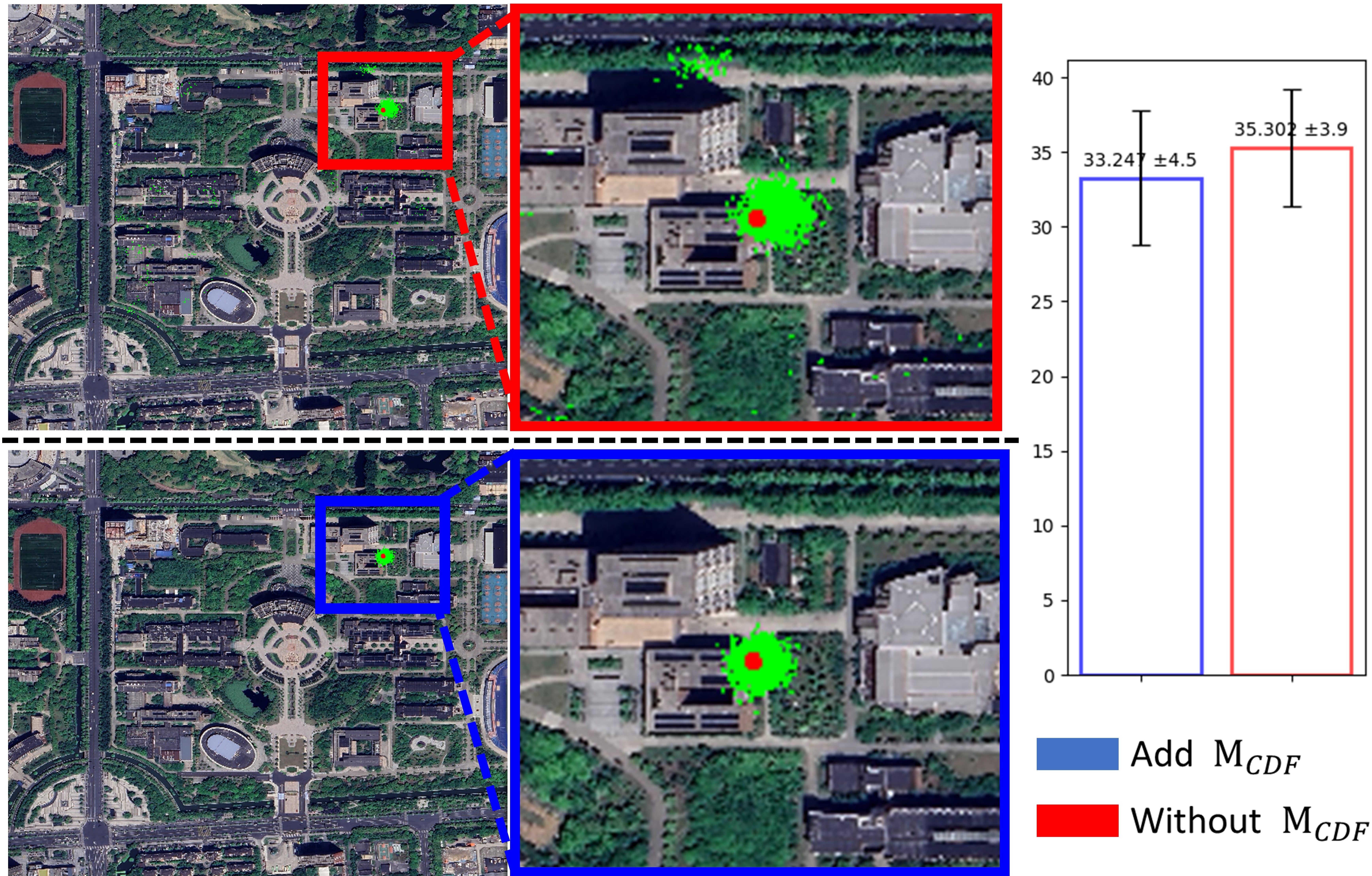}
    \caption{Comparison of performance and fitting rate with \( M_{CDF} \). Particle state diagrams are recorded with and without the \( M_{CDF} \) method, showing improved fitting rate and central error after incorporating the \( M_{CDF} \) method.}
    \label{fig14}
\end{figure}

We observe that the central image weight influences alignment. To improve this, we incorporate the central distance field into the particle weight calculation. Testing the performance of MAFS-10 with and without the central distance field (Fig. \hyperref[fig14]{15}) under identical parameters shows that the average particle fitting rate improves by 5.8\% with the addition of the central distance field. Furthermore, comparing the scattering intervals of the particles post-fitting (Fig. \hyperref[fig14]{15}) demonstrates that adding the central distance field effectively reduces particle scattering.

\subsubsection{Semantic Weight Allocation}

We analyze the weight distribution for each semantic category. Localization methods based on a single semantic fail consistently. When testing localization methods that combine two semantics (e.g., road category with any other category), all combinations fail due to insufficient feature information, particularly because most roads in the test scenarios are located around the perimeter. In contrast, all combinations of the four semantics complete the fitting, though at a slower rate than when all four semantics are used.

\begin{table*}[!t]
    \caption{Comparison with Existing State-of-the-Art Methods} 
    \label{tab3}
    \resizebox{\textwidth}{!}{
    \begin{tabular}{lccccc}
    \hline
    \textbf{Method} & \textbf{Map Size (\(m^{2}\))} & \textbf{Map Size (\( p^{2} \))}  & \textbf{Trajectory (m)} & \textbf{Error (m)} & \textbf{Error (p)}\\
    \hline
    UAVD4L \cite{UAVD4L}          & 2425 × 1054 & -      & -           & \textless 5  & -        \\
    abBRIEF \cite{abb}            & 1160 × 1160 & 4800 × 4800 & 2400         & 17.78       & 74.08     \\
    Masselli et al. \cite{bib51}  & 150 × 90    & 3440 × 3440 & \textless 500 & 12.75       & 63.7      \\
    Hong et al. \cite{bib35}      & -           & -           & 600          & \textless 10 & -        \\
    \hline
    SA-PF (ours)                 & 2000 × 2000 & 7615 × 5258 & 4000         & \textless 9  & \textless 35 \\
    \hline
    \end{tabular}}
\end{table*}

\subsection{Comparison with the Existing State-of-the-Art}
\label{section IVC}

Section \hyperref[section II]{2} discusses satellite imagery-based UAV localization systems most comparable to our approach, including works by Masselli et al. \cite{bib51}, Mantelli et al. \cite{abb}, Wu et al. \cite{UAVD4L}, Dai et al. \cite{UL14}, and Hong et al. \cite{bib35}. Table \hyperref[tab3]{III} provides a systematic comparison of methodological parameters, with limited analysis due to unavailable datasets or codes from benchmark studies.

A critical evaluation metric for satellite-based localization is the image-to-ground scale factor (pixel-to-meter resolution ratio). Larger scale factors increase per-pixel distance representation, amplifying positioning errors and algorithmic precision requirements. Our dataset exhibits superior scale characteristics, being 6.02× smaller than \cite{bib51} and 1.08× smaller than \cite{abb}.

Our system achieves a localization accuracy of under 9 meters. In contrast to the deep learning-based FPI method, which requires pre-training for each new scenario and operates within constrained search parameters, our approach allows for localization across substantially larger spatial dimensions. The three-dimensional search space (width, height, altitude) supported by our system exceeds conventional methods by three orders of magnitude. While matching the precision of the \cite{UAVD4L} method, our solution reduces environmental dependency requirements, requiring only a single low-resolution 2D satellite image as input. The UAVD4L system demonstrates 6-DoF localization in GPS-denied environments via hybrid offline-online processing, but its reliance on detailed 3D textured models imposes significant pre-mapping requirements, limiting operational flexibility.

Our system achieves positioning errors under 9m while overcoming critical limitations of existing approaches. Unlike the deep learning-based method \cite{UL14}, which requires scene-specific pretraining and constrained search spaces, we enable localization across a 1000× larger 3D search volume. While matching the accuracy of Hong et al. \cite{bib35} and Wu et al. \cite{UAVD4L}, our approach reduces prior information requirements to a single low-resolution 2D satellite image, contrasting with UAVD4L’s computationally intensive 6-DoF localization framework that demands detailed 3D textured models.

We further evaluate performance through the normalized error-to-map ratio (EMR: average error/map dimension). Lower EMR values indicate superior scalability, with our system achieving a benchmark 4.5 ‰ ratio, which is 3.4× and 17.7× lower than \cite{abb} and \cite{bib51}, respectively.

\section{Conclusion}

This study addresses the limitations of traditional retrieval-based UAV localization approaches, which are often hindered by dataset constraints, suboptimal real-time performance, environmental sensitivity, and poor generalization in dynamic or temporally varying conditions. We introduce the MAFS dataset, a large-scale, multi-altitude flight segment collection designed to provide high-quality data for UAV localization research across diverse urban environments and variable altitude scenarios. Building on this dataset, we propose a novel and efficient approach, the Semantic-Weighted Adaptive Particle Filter (SWA-PF). By integrating robust semantic features from both UAV-captured images and satellite imagery via a semantic weighting mechanism and optimizing the particle filtering architecture, our method achieves a tenfold increase in computational efficiency compared to conventional feature extraction techniques. Additionally, it maintains global positioning errors below 10 meters and enables rapid 4-DoF pose estimation within seconds using low-resolution satellite maps. These results demonstrate the efficacy of SWA-PF for reliable UAV navigation in dynamic, real-world environments.

\section{Limitations}

Despite the significant advancements presented, our method has certain limitations. One major challenge arises in scenarios with limited distinctive visual features, such as when large portions of the field of view are dominated by buildings or vegetation. In such cases, the method struggles to accurately estimate the UAV's pose due to the lack of sufficient reference points for spatial alignment. Additionally, while our approach performs well under various conditions, it is sensitive to perspective discrepancies between oblique satellite images and UAV top-view images. These geometric mismatches lead to amplified positioning errors.

\section{Future Work}

Future research will focus on methods to estimate the UAV or gimbal camera roll/pitch angles, thereby extending the positioning capability to 6-DoF, which would provide a more complete representation of the UAV's orientation in space. We also plan to explore the application of our method at higher altitudes, where increased viewpoint variability and complexity present new challenges. Furthermore, we aim to continue refining our semantic weighting mechanism and particle filtering architecture to enhance performance in feature-scarce environments and reduce sensitivity to perspective disparities.

\bibliographystyle{IEEEtran}
\bibliography{IEEEabrv,SWAPF}
\end{document}